\documentclass{article}

\usepackage[final, dblblindworkshop]{coml_2025}

\usepackage{amsmath}
\usepackage{amssymb}
\usepackage{mathtools}

\usepackage{amsthm}
\workshoptitle{Constrained Optimization for Machine Learning (COML)}

\usepackage[utf8]{inputenc} 
\usepackage[T1]{fontenc}    
\usepackage{hyperref}       
\usepackage{url}            
\usepackage{booktabs}       
\usepackage{amsfonts}       
\usepackage{nicefrac}       
\usepackage{microtype}      
\usepackage{xcolor}         

\usepackage[utf8]{inputenc}
\usepackage[english]{babel}
\usepackage[T1]{fontenc}
\usepackage{algorithm}
\usepackage{algpseudocode}
\usepackage{url}
\usepackage{enumitem}
\usepackage{thmtools}
\usepackage{amsthm}
\usepackage{amsmath}
\usepackage{amssymb}
\usepackage{mathrsfs}
\usepackage{geometry}
\usepackage{svg}
\usepackage{graphicx}
\usepackage{titlesec}
\usepackage{hyphenat}
\usepackage{comment}
\usepackage{tabularx}
\usepackage{yfonts}
\usepackage{acronym}
\usepackage{wrapfig}

\usepackage[export]{adjustbox}
\usepackage{appendix}
\usepackage{hyperref}
\usepackage{amsthm}
\usepackage{amsmath}
\usepackage{amssymb}

\author{
Fatmazohra Rezkellah  \\
  {{ Department of Computer Science,}} \\ {{Université Paris-Dauphine, France }}
 \\
  \texttt{fatma-zohra.rezkellah@dauphine.eu}
\And
   Ramzi Dakhmouche\thanks{Corresponding author.}  \\
{  {Institute of Mathematics, EPFL, Switzerland }}\\
 {  Computational Engineering Lab, Empa, Switzerland} \\
 \texttt{ramzi.dakhmouche@epfl.ch} \\
}

\title{Machine Unlearning Meets Adversarial Robustness via Constrained Interventions on LLMs}

\begin{document}

\maketitle

\begin{abstract}
With the increasing adoption of Large Language Models (LLMs), more customization is needed to ensure privacy-preserving and safe generation. We address this objective from two critical aspects: unlearning of sensitive information and robustness to jail-breaking attacks. We investigate various constrained optimization formulations that address both aspects in a \emph{unified manner}, by finding the smallest possible interventions on LLM weights that either make a given vocabulary set unreachable or embed the LLM with robustness to tailored attacks by shifting part of the weights to a \emph{safer} region. Beyond unifying two key properties, this approach contrasts with previous work in that it doesn't require an oracle classifier that is typically not available or represents a computational overhead. Surprisingly, we find that the simplest point-wise constraint-based intervention we propose leads to better performance than max-min interventions, while having a lower computational cost. Comparison against state-of-the-art defense methods demonstrates superior performance of the proposed approach. 
\end{abstract}

\section{Introduction}

Ensuring the safety of generative model outputs is a critical requirement for their deployment in real-world and safety-sensitive applications such as online content moderation and confidential data processing. To mitigate the risks of generating toxic or harmful content, prior work has explored strategies such as fine-tuning \cite{krause2021gedi, dengresidual}, prompt engineering \cite{luo2023prompt, xie2023defending}, uncertainty quantification \cite{dakhmouche2025can} and model augmentation \cite{dai2022knowledge, meng2022locating}. However, fine-tuning and model augmentation are computationally expensive, while prompt-based defenses are often brittle and sensitive to changes in model, context, and adversarial manipulation.

A more efficient alternative leverages lightweight probes (e.g., classifiers) on the activation space to steer generations toward desired attributes such as topic or sentiment \cite{dathathriplug, subramani2022extracting, hernandezinspecting, konen2024style, cheng2024linearly}. While effective at reducing computational overhead, these methods remain heavily constrained by the probe’s limited training data and are ineffective in defending against adversarial attacks. Recent works have instead turned to adversarial detection, analyzing statistical signatures of perturbed inputs or comparing them against safe baselines \cite{candogansingle2025, robeysmoothllm2025}.

Conversely, the task of unlearning sensitive or confidential content has mainly been addressed through partial retraining in teacher–student frameworks \cite{chen2023unlearn} or iterative fine-tuning \cite{yao2024large, tian2024forget}, both of which incur substantial computational cost. Thus, there remains a gap for methods that jointly address robustness and unlearning while remaining computationally efficient. In this work, inspired by principled robustness approaches for regression \cite{dakhmouche2025noise, dakhmouche2025network, dakhmoucherobust}, we bridge this gap by introducing a simple yet effective approach that simultaneously tackles adversarial robustness and unlearning. Specifically, our contributions can be summarized as follows: 
\begin{itemize}
    \item We propose and solve various constrained optimization problems to embed LLMs with robustness to adversarial attacks, while allowing them to unlearn unwanted content,
    \item We overcome the need for artificial probes by introducing a continuous relaxation over the space of prompts and performing interventions that directly constrain concept embeddings,
    \item We demonstrate a gain in performance in comparison to state-of-the-art defense algorithms and set a state-of-the-art for affordable unlearning on LLMs.
\end{itemize}

\section{Problem Setting}
Given a trained language model $\ell: \Sigma \longrightarrow \Sigma$, we consider two fundamental safety-related tasks: 
\begin{enumerate}
    \item How to delete certain information, instantiated as a given set of words, from the generation space of $\ell$ at a minimal computational cost ?
    \item How to make $\ell$ more robust to jail-breaking adversarial attacks that lead to dangerous or toxic content ?
\end{enumerate}
To address both questions of unlearning and robustness in a unified way, we leverage the fact that information is implicitly stored in latent space pathways, hence constraining them by partially steering weights simultaneously removes both sensitive and toxic content. To achieve that, several strategies warrant attention depending on whether one chooses to maximize distance from risky regions, minimize distance from safe ones or target specific concepts in a point-wise manner. We present in the next section the formal constrained optimization problem descriptions for each of these strategies and discuss how to solve them. 
\section{Constrained Model Interventions}
We present in the following three constrained intervention approaches we propose, as well as the related strengths and drawbacks of each approach. 
\subsection{Towards Safer Regions (TSR)}
In the first formulation, we seek to find minimal weight perturbations that maximize the probability of safe responses to jail-breaking prompts

\begin{equation}
\min_{\|{\delta}\| \leq \varepsilon} \quad \mathcal{L}_{\textrm{safety}}(\ell_{\boldsymbol{\theta} + \boldsymbol{\delta}}(\mathbf{x}), \mathbf{y}_{\textrm{safe}}), \\
\end{equation}

where $\mathcal{L}_{\textrm{safety}}$ represents the safety-oriented loss function that encourages the generation of appropriate refusal responses, $\theta  \in \mathbb{R}^d$ the model parameters, $\mathbf{y}_{\textrm{safe}}$ the target safe outputs, $x$ a given jailbreaking input embedding and $\varepsilon > 0$ the perturbation budget that constrains the magnitude of weight modifications. 

The safety loss function is defined as:
\begin{equation}
\label{eq:safety_loss}
\mathcal{L}_{\textrm{safety}}(f_{\boldsymbol{\theta} + \boldsymbol{\delta}}(\mathbf{x}), \mathbf{y}_{\textrm{safe}}) = -\log\left(\sum_{k \in \mathcal{K}_{\textrm{safety}}} p_k(\mathbf{x}; \boldsymbol{\theta} + \boldsymbol{\delta})\right),
\end{equation}

where $\mathcal{K}_{\textrm{safety}}$ is the set of token indices corresponding to safety keywords (e.g., "I cannot", "I apologize", "inappropriate"...), and $p_k(\mathbf{x}; \boldsymbol{\theta} + \boldsymbol{\delta})$ represents the probability assigned to token $k$ by the perturbed model, for input prompt $x$. Among the advantages of this formulation is that, one does not need examples of dangerous generation, which requires annotation. Furthermore, it can simply be solved via projected-gradient descent. However, the model output embeddings are constrained towards safe generation in a soft way, which leads to weaker performance compared to the hard constraint formulation presented in subsection 3.3 below. 

\subsection{Away from Risky Regions (ARR)}

Given the popularity min-max training in adversarial defense on images \cite{madry2018towards, robeyadversarial2024}, we explore such a formulation in the context of LLMs. Specifically, we consider the max-min problem

\begin{equation}
\label{eq:formulation2_constrained}
\begin{aligned}
\max_{\|\boldsymbol{\delta} \| \leq \varepsilon} \,\, \min_{\mathbf{x} \in \mathcal{X}} \quad &\mathcal{L}_{\text{harmful}}(\ell_{\boldsymbol{\theta} + \boldsymbol{\delta}}(\mathbf{x}), \mathbf{y}_{\text{harmful}}). 
\end{aligned}
\end{equation}

This formulation seeks weight perturbations that minimize the likelihood of generating harmful content, under worst-case input scenarios $x \in \mathcal{X}$, where $\mathcal{X}$ is the prompt space. The harmful loss function is defined as:
\begin{equation}
\label{eq:harmful_loss}
\mathcal{L}_{\text{harmful}}(\ell_{\boldsymbol{\theta} + \boldsymbol{\delta}}(\mathbf{x}), \mathbf{y}_{\text{harmful}}) = -\log\left(\sum_{k \in \mathcal{K}_{\text{harmful}}} p_k(\mathbf{x}; \boldsymbol{\theta} + \boldsymbol{\delta})\right), 
\end{equation}

where $\mathcal{K}_{\text{harmful}}$ contains token indices associated with harmful instruction-following behavior, and $x \in \mathcal{X}$ is an input.
Given the discrete structure of $\mathcal{X}$, one needs a continuous relaxation to solve problem (\ref{eq:formulation2_constrained}) efficiently. For that matter, use a probabilistic relaxation over the number of tokens, leading to an optimization over the simplex which can be solved via projected gradient descent \cite{geisler2024attacking}. Although this approach might be more principled, it requires knowledge or choice of a set of harmful concepts and might be more conservative, given that it considers worst-case prompts.  
\subsection{Point-Wise Constrained Regions (PCR)}
Last, we explore a simple point-wise constraint strategy that is based on finding the smallest intervention that makes the LLM MLP activations, for a jailbreaking prompt, non-equal to embedding of dangerous output. More precisely,  denoting by $\mathcal{C} = \{ c_1, \dots, c_n \}$  a set of forbidden concept embeddings and by $o^{(l)} (x, \theta)$ the output of layer $l \leq L$ corresponding to input prompt $x$ and parameters $\theta$, the goal is to solve for each $l \leq L$, 
\begin{equation}
\begin{aligned}
\min_{\theta^{(l)} \in \mathbb{R}^{d_l}} &\quad \|\delta^{l}\|_2^2 \\
\text{subject to} &\quad \|o^{(l)}(x; \theta + \delta^{l} ), c_i \|_2 \geq \varepsilon, \,\, \forall i \le n \\
\end{aligned} \, 
\end{equation}
where we abuse notation to mean update in the parameters corresponding to the $l$-th layer of the MLP when writing $\theta + \delta^{l}$. The solution can then be written under closed form based on the Karush–Kuhn–Tucker (KKT) conditions. We report it in Appendix C. Similar to the previous strategy, the main drawback of this formulation is the fact that it requires a set of forbidden concepts $\mathbf{C}$. Yet, given its simplicity, it allows for a semi-closed-form solution which, surprisingly, leads to the best performance even exceeding state-of-the-art defense approaches, as illustrated in section 4. 
\section{Experimental Results}
We evaluate the various intervention strategies on a custom obedience dataset (reported in Appendix B) proposed by \cite{geisler2024attacking} as well as on the standard LLM defense benchmark HarmBench \cite{mazeika2024harmbench}. For robustness evaluation, we compare against state-of-the-art adversarial defense algorithms SmoothLLM \cite{robeysmoothllm2025} and Self-reminder \cite{xie2023defending}. We measure performance via adversarial attack success rate (ASR) as well as refusal levels (see appendix A for details). We compute the attacks using the state-of-the-art LLM jailbreaking algorithm proposed by \cite{geisler2024attacking}. We report running times in Appendix A. As for unlearning levels, we measure them by comparing perplexity levels of given sequences before and post-intervention. Since perplexity is a proxy for how likely a sequence is, this represents a practical measure for unlearning levels. Due to the lack of affordable unlearning algorithms for LLMs, we simply compare the models against themselves post-intervention. 

\subsection{Adversarial Robustness }
The results shown in Tables \ref{tab:defense_comparison_harmbench} and \ref{tab:refusal_analysis} demonstrate that our point-wise intervention approach has higher performance than the competitive approaches. As for Table \ref{tab:attack_success_rate-base}, it shows that formulations 1 and 2 barely improve over the unprotected baseline. 
\begin{table}[H]
\caption{Attack Success Rate with different defenses - HarmBench Dataset- Lower better}
\label{tab:defense_comparison_harmbench}
\centering
\begin{tabular}{|l|c|c|c|ccc|}
\hline
\textbf{Model} & \textbf{No Defense} & \textbf{PointWise (Ours)} & \textbf{SmoothLLM} & \multicolumn{3}{c|}{\textbf{Self-Reminder}} \\
\cline{5-7}
 &  &  &  & \textbf{Basic} & \textbf{Warn} & \textbf{Praise} \\
\hline
Llama-3.1 8B     & 11.0 & \textbf{0.0}   & 7.245 & 2.03 & 1.98 & 0.8 \\
\hline
Mistral 7B v0.2  & 30.0 & \textbf{5.88}  & 18.9  & 23.1 & 19.8 & 28.5 \\
\hline
Gemma 2B-IT      & 22.0 & \textbf{2.508} & 8.225 & 12.98 & 12.76 & 19.58 \\
\hline
\end{tabular}
\end{table}

\begin{table}[H]
\caption{Refusal Pattern Analysis in Protected Completions - Higher better}
\label{tab:refusal_analysis}
\centering
\begin{tabular}{|l|c|c|c|}
\hline
\textbf{Model} & \textbf{Our intervention (\%)} & \textbf{SmoothLLM (\%)} & \textbf{Self Reminder (\%)} \\
\hline
Llama-3.1 8B Instruct & \textbf{100.0} & 87.5 & 24.3 \\
\hline
Gemma 2B-IT & \textbf{97.4} & 10 & 36.9 \\
\hline
Mistral 7B v0.2 & 26.7 & 37.5 & 20 \\
\hline
\end{tabular}
\end{table}
\begin{table*}[ht]
\centering
\caption{Attack Success Rate with different defenses - Custom Dataset- Lower better}
\label{tab:attack_success_rate-base}

\begin{tabular}{|l|c|c|c|ccc|}
\hline
\textbf{Model} & \textbf{No Defense} & \textbf{PointWise (Ours)} & \textbf{SmoothLLM} & \multicolumn{3}{c|}{\textbf{Self-Reminder}} \\
\cline{5-7}
 &  &  &  & \textbf{Basic} & \textbf{Warn} & \textbf{Praise} \\
\hline
Llama-3.1 8B     & 60.50 & \textbf{5.49} & 30.17 & 3.00 & 2.00 & 18.00 \\
\hline
Mistral 7B v0.2  & 90.00 & 36.47 & 61.00 & 77.00 & 66.00 & 95.00 \\
\hline
Gemma 2B-IT      & 85.30 & 30.50 & 56.20 & 59.00 & 58.00 & 89.00 \\
\hline
\end{tabular}
\end{table*}

\subsection{ Machine Unlearning}
Once again, the numerical results reported in Table \ref{tab:perplexity_results} show the effectiveness of the point-wise intervention strategy in unlearning given sets of words or concepts at a much reduced computational cost. 
\begin{table}[H]
\centering
\caption{Forbidden Word Perplexity Analysis: Baseline vs. Intervention Results}
\label{tab:perplexity_results}
\begin{tabular}{|l|l|c|c|c|c|}
\hline
Model & Dataset & Baseline & PointWise Intervention  \\
\hline
Gemma-2B-IT & Obedience & 8.816 & 12.72  \\
Gemma-2B-IT & HarmBench & 16.757 & 18.157  \\
\hline
Mistral-7B-Instruct & Obedience & 8.627 & 13.74  \\
Mistral-7B-Instruct & HarmBench & 6.195 & 8.005  \\
\hline
Llama-3-8B-Instruct & Obedience & 6.48 & 7.735  \\
Llama-3-8B-Instruct & HarmBench & 10.08 & 13  \\
\hline
\end{tabular}
\end{table}

\section{Conclusion}
In this work, we introduced a constrained optimization framework that jointly addresses the challenges of unlearning and robustness in large language models, enabling more reliable and effective model customization. By formulating these objectives leveraging relaxations over the space of prompts and tailored point-wise constraints, our approach reduces reliance on external probes, which are often brittle and require substantial computational resources to train at scale. This framework provides a lightweight yet principled alternative, paving the way for safer and more efficient deployment of generative models in real-world settings, especially for smaller organizations and communities with limited resources. 

\section*{Broader Impact Statement}
This work advances the safety of generative models by introducing a computationally efficient framework for robustness and unlearning. In principle, such methods can reduce the risk of harmful outputs (e.g., toxic, biased, or confidential content) and thereby enable safer integration of generative AI into sensitive domains such as education, healthcare, and online platforms. At the same time, more powerful unlearning techniques could be misapplied—for instance, to obscure accountability or selectively erase undesirable information. Careful governance and responsible deployment will therefore be essential to ensure that these methods are used to promote safety and accessibility without introducing new harms.

\section*{Acknowledgements}
This work was supported by the Swiss National Science Foundation under grant No. 212876. We acknowledge computational resources from the Swiss National Supercomputing Centre CSCS. The authors acknowledge Prof. Hossein Gorji for valuable discussions and insightful feedback. R.D. acknowledges Dr. Ivan Lunati for providing laboratory infrastructure and computational resources.

\bibliography{Biblio}

\begin{thebibliography}{10}

\bibitem{candogansingle2025}
Leyla~Naz Candogan, Yongtao Wu, Elias~Abad Rocamora, Grigorios Chrysos, and Volkan Cevher.
\newblock Single-pass detection of jailbreaking input in large language models.
\newblock {\em Transactions on Machine Learning Research}, 2025.

\bibitem{chen2023unlearn}
Jiaao Chen and Diyi Yang.
\newblock Unlearn what you want to forget: Efficient unlearning for llms.
\newblock In {\em Proceedings of the 2023 Conference on Empirical Methods in Natural Language Processing}, pages 12041--12052, 2023.

\bibitem{cheng2024linearly}
Emily Cheng, Marco Baroni, and Carmen~Amo Alonso.
\newblock Linearly controlled language generation with performative guarantees.
\newblock {\em arXiv preprint arXiv:2405.15454}, 2024.

\bibitem{dai2022knowledge}
Damai Dai, Li~Dong, Yaru Hao, Zhifang Sui, Baobao Chang, and Furu Wei.
\newblock Knowledge neurons in pretrained transformers.
\newblock In {\em Proceedings of the 60th Annual Meeting of the Association for Computational Linguistics (Volume 1: Long Papers)}, pages 8493--8502, 2022.

\bibitem{dakhmouche2025can}
Ramzi Dakhmouche, Adrien Letellier, and Hossein Gorji.
\newblock Can linear probes measure llm uncertainty?
\newblock {\em arXiv preprint arXiv:2510.04108}, 2025.

\bibitem{dakhmouche2025network}
Ramzi Dakhmouche, Ivan Lunati, and Hossein Gorji.
\newblock Network system forecasting despite topology perturbation.
\newblock In {\em ICML 2025 Workshop on Scaling Up Intervention Models}, 2025.

\bibitem{dakhmouche2025noise}
Ramzi Dakhmouche, Ivan Lunati, and Hossein Gorji.
\newblock Noise tolerance of distributionally robust learning.
\newblock In {\em ICML 2025 Workshop on Scaling Up Intervention Models}, 2025.

\bibitem{dakhmoucherobust}
Ramzi Dakhmouche, Ivan Lunati, and Hossein Gorji.
\newblock Robust symbolic regression for dynamical system identification.
\newblock {\em Transactions on Machine Learning Research}, 2025.

\bibitem{dathathriplug}
Sumanth Dathathri, Andrea Madotto, Janice Lan, Jane Hung, Eric Frank, Piero Molino, Jason Yosinski, and Rosanne Liu.
\newblock Plug and play language models: A simple approach to controlled text generation.
\newblock In {\em International Conference on Learning Representations}, 2020.

\bibitem{dengresidual}
Yuntian Deng, Anton Bakhtin, Myle Ott, Arthur Szlam, and Marc'Aurelio Ranzato.
\newblock Residual energy-based models for text generation.
\newblock In {\em International Conference on Learning Representations}, 2020.

\bibitem{geisler2024attacking}
Simon Geisler, Tom Wollschl{\"a}ger, MHI Abdalla, Johannes Gasteiger, and Stephan G{\"u}nnemann.
\newblock Attacking large language models with projected gradient descent.
\newblock In {\em ICML 2024 Next Generation of AI Safety Workshop}.

\bibitem{hernandezinspecting}
Evan Hernandez, Belinda~Z Li, and Jacob Andreas.
\newblock Inspecting and editing knowledge representations in language models.
\newblock In {\em First Conference on Language Modeling}, 2024.

\bibitem{konen2024style}
Kai Konen, Sophie Jentzsch, Diaoul{\'e} Diallo, Peer Sch{\"u}tt, Oliver Bensch, Roxanne El~Baff, Dominik Opitz, and Tobias Hecking.
\newblock Style vectors for steering generative large language models.
\newblock In {\em Findings of the Association for Computational Linguistics: EACL 2024}, pages 782--802, 2024.

\bibitem{krause2021gedi}
Ben Krause, Akhilesh~Deepak Gotmare, Bryan McCann, Nitish~Shirish Keskar, Shafiq Joty, Richard Socher, and Nazneen~Fatema Rajani.
\newblock Gedi: Generative discriminator guided sequence generation.
\newblock In {\em Findings of the Association for Computational Linguistics: EMNLP 2021}, pages 4929--4952, 2021.

\bibitem{luo2023prompt}
Yifan Luo, Yiming Tang, Chengfeng Shen, Zhennan Zhou, and Bin Dong.
\newblock Prompt engineering through the lens of optimal control.
\newblock {\em arXiv preprint arXiv:2310.14201}, 2023.

\bibitem{madry2018towards}
Aleksander Madry, Aleksandar Makelov, Ludwig Schmidt, Dimitris Tsipras, and Adrian Vladu.
\newblock Towards deep learning models resistant to adversarial attacks.
\newblock In {\em International Conference on Learning Representations}, 2018.

\bibitem{mazeika2024harmbench}
Mantas Mazeika, Long Phan, Xuwang Yin, Andy Zou, Zifan Wang, Norman Mu, Elham Sakhaee, Nathaniel Li, Steven Basart, Bo~Li, et~al.
\newblock Harmbench: A standardized evaluation framework for automated red teaming and robust refusal.
\newblock In {\em International Conference on Machine Learning}, pages 35181--35224. PMLR, 2024.

\bibitem{meng2022locating}
Kevin Meng, David Bau, Alex Andonian, and Yonatan Belinkov.
\newblock Locating and editing factual associations in gpt.
\newblock {\em Advances in neural information processing systems}, 35:17359--17372, 2022.

\bibitem{robeyadversarial2024}
Alexander Robey, Fabian Latorre, George~J Pappas, Hamed Hassani, and Volkan Cevher.
\newblock Adversarial training should be cast as a non-zero-sum game.
\newblock In {\em The Twelfth International Conference on Learning Representations}, 2024.

\bibitem{robeysmoothllm2025}
Alexander Robey, Eric Wong, Hamed Hassani, and George~J Pappas.
\newblock Smoothllm: Defending large language models against jailbreaking attacks.
\newblock {\em Transactions on Machine Learning Research}, 2025.

\bibitem{subramani2022extracting}
Nishant Subramani, Nivedita Suresh, and Matthew~E Peters.
\newblock Extracting latent steering vectors from pretrained language models.
\newblock In {\em ACL (Findings)}, 2022.

\bibitem{tian2024forget}
Bozhong Tian, Xiaozhuan Liang, Siyuan Cheng, Qingbin Liu, Mengru Wang, Dianbo Sui, Xi~Chen, Huajun Chen, and Ningyu Zhang.
\newblock To forget or not? towards practical knowledge unlearning for large language models.
\newblock In {\em Findings of the Association for Computational Linguistics: EMNLP 2024}, pages 1524--1537, 2024.

\bibitem{xie2023defending}
Yueqi Xie, Jingwei Yi, Jiawei Shao, Justin Curl, Lingjuan Lyu, Qifeng Chen, Xing Xie, and Fangzhao Wu.
\newblock Defending chatgpt against jailbreak attack via self-reminders.
\newblock {\em Nature Machine Intelligence}, 5(12):1486--1496, 2023.

\bibitem{yao2024large}
Yuanshun Yao, Xiaojun Xu, and Yang Liu.
\newblock Large language model unlearning.
\newblock {\em Advances in Neural Information Processing Systems}, 37:105425--105475, 2024.

\end{thebibliography}
\bibliographystyle{plain}

\newpage

\appendix
\section{Additional Numerical Results \& Experimental Setup}
\begin{figure}[H]
    \centering
    \includegraphics[width=10cm]{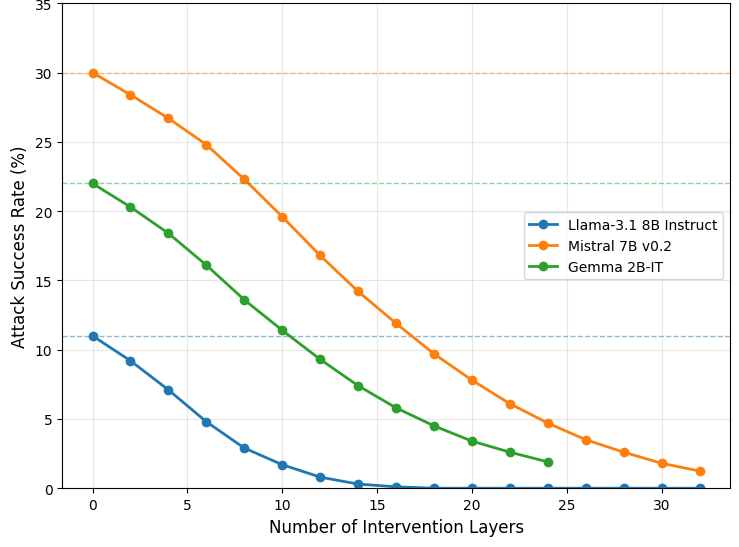}
    \caption{Intervention Configuration Comparison - HarmBench Dataset}
    \label{fig:my_label1}
\end{figure}

\begin{figure}[H]
    \centering
        \includegraphics[width=10cm]{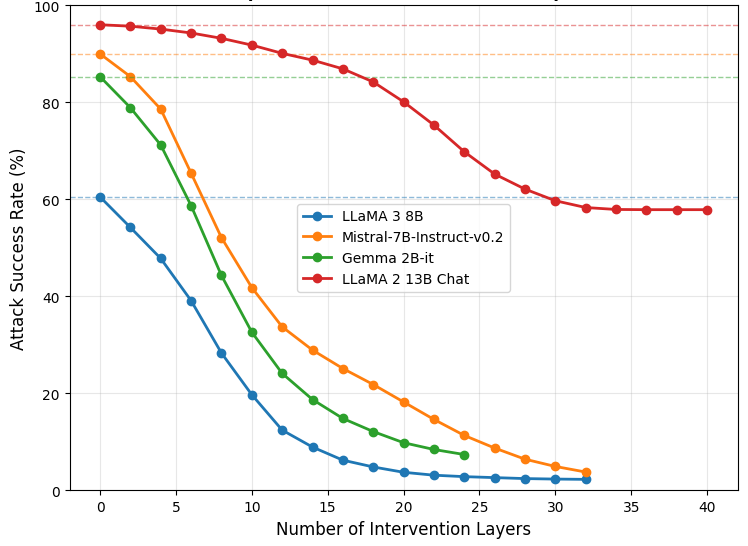}
    \caption{Intervention Configuration Comparison - Custom Obedience Dataset}
    \label{fig:my_label2}
\end{figure}

\textbf{Performance Measures.} Attack success rate only captures whether the exact word or sentence have been generated. As for refusal levels, they measure when the model completely refuses to respond, making it a stronger measure of performance. \\ \ \\
\textbf{Performance Layer Scaling.} We report in Figures \ref{fig:my_label1} and \ref{fig:my_label2} the evolution of performance as more MLP layers used in interventions. \\ \ \\
\textbf{HarmBench Dataset - Intervention Scope Analysis:}

\textbf{Running Times.} 

\begin{table}[H]
\caption{Defense Method Timing Comparison (Average per test case)}
\label{tab:timing_analysis}
\centering
\begin{tabular}{|l|c|c|c|}
\hline
\textbf{Model} & \textbf{Attack Time (s)} & \textbf{Our PCR approach (s)} & \textbf{SmoothLLM(s)} \\
\hline
LLaMA 3 8B & 38.89 & \textbf{20.16} & 36.12 \\
\hline
Mistral-7B-Instruct-v0.2 & 27.43 & \textbf{17.28} & 40.17 \\
\hline
Gemma 2B-it & 14.375 & \textbf{10.44} & 11.62 \\
\hline
\end{tabular}
\end{table}
\textit{Note}: Self-Reminder execution times were not recorded as this method introduces negligible computational overhead. The approach operates solely through augmentation of the system prompt with reminder text, requiring no additional forward passes or model modifications during inference.
\section{Obedience Dataset}
We generate the following forbidden keywords as a complementary dataset: "abuse", "accused", "aggressive", "annihilate", "artillery", "assassinate", "assault", "attack", "battle", "blade", "bleed", "blood", "bomb", "brutality", "cannon", "carnage", "chaos", "combat", "conflict", "crime", "criminal", "cruelty", "culprit", "damage", "deadly", "death", "defendant", "demolish", "destroy", "detonation", "discipline", "disorder", "dynamite", "eliminate", "execute", "explosion", "explosive", "fatal", "fear", "felon", "fight", "frighten", "gore", "grenade", "guilty", "gun", "harm", "horror", "hurt", "injure", "intimidate", "invasion", "judgment", "justice", "kill", "knife", "massacre", "mayhem", "menace", "merciless", "missile", "murder", "nightmare", "obliterate", "offender", "panic", "penalty", "perpetrator", "pistol", "poison", "punishment", "raid", "rebellion", "retaliation", "revenge", "revolt", "rifle", "rocket", "ruthless", "savage", "scare", "sentence", "shoot", "slash", "slaughter", "stab", "suspect", "sword", "terror", "terrorize", "threat", "torture", "toxic", "vengeance", "vicious", "violence", "violent", "war", "warfare", "wound". 

\section{Closed-form Solution for Point-Wise Interventions}
We write down the KKT conditions for the third formulation of the robustness problem. Recall
$$
\begin{aligned}
\min_{\delta \in \mathbb{R}^{d_l}} \quad & \|\delta\|_2^2 \\
\text{s.t.} \quad & \|\,o^{(l)}(x;\theta+\delta) - c_i\,\|_2 \;\;\geq\; \varepsilon, 
\quad i=1,\dots,n.
\end{aligned}
$$
The Lagrangian is given by: 
$$
\mathcal{L}(\delta,\lambda) 
= \|\delta\|_2^2 + \sum_{i=1}^n \lambda_i\Big( \varepsilon - \|\,o^{(l)}(x;\theta+\delta)-c_i\,\|_2 \Big),
\quad \lambda_i \ge 0.
$$
Leading to KKT Conditions
$$
\begin{aligned}
&\text{(Primal feasibility)} && \|\,o^{(l)}(x;\theta+\delta^\star) - c_i\,\|_2 \;\;\geq\; \varepsilon, 
& i=1,\dots,n, \\[6pt]
&\text{(Dual feasibility)} && \lambda_i^\star \;\;\geq\; 0, 
& i=1,\dots,n, \\[6pt]
&\text{(Complementary slackness)} && 
\lambda_i^\star \Big(\varepsilon - \|\,o^{(l)}(x;\theta+\delta^\star)-c_i\,\|_2\Big)=0, 
& i=1,\dots,n, \\[6pt]
&\text{(Stationarity)} && 
\nabla_\delta \mathcal{L}(\delta^\star,\lambda^\star) = 
2\delta^\star - \sum_{i=1}^n \lambda_i^\star \, J_i(\delta^\star)^\top 
\frac{o^{(l)}(x;\theta+\delta^\star)-c_i}{\|\,o^{(l)}(x;\theta+\delta^\star)-c_i\,\|_2}
= 0,
\end{aligned}
$$
where 
$$
J_i(\delta) := \frac{\partial o^{(l)}(x;\theta+\delta)}{\partial \theta^{(l)}}
$$
\textbf{Single-Constraint Building Block}: For a single forbidden concept, the closed-form solution serves as the fundamental building block:
\begin{equation}
\delta^{(l)*}_{\text{single}} = \frac{[\epsilon - \|r_i\|_2]_+}{\|h_{\text{intermediate}}\|_2^2} \frac{r_i h_{\text{intermediate}}^T}{\|r_i\|_2}
\end{equation}
where $[\cdot]_+ = \max(0, \cdot)$ denotes the positive part function and $h_{\text{intermediate}}$ is the output of previous layer $l-1$. 

\textbf{Most Violated Constraint Approach}: For the multi-constraint case, we employ a practical approach based on constraint violation priority. The violation severity for each concept is measured as:
\begin{equation}
\text{violation}_i = \max(0, \epsilon - \|r_i\|_2)
\end{equation}

The most violated constraint is identified as:
\begin{equation}
j^* = \arg\max_i \text{violation}_i = \arg\min_i \|r_i\|_2 \quad \text{(among violated constraints)}
\end{equation}

\textbf{Iterative Multi-Constraint Solution}: The complete solution employs an iterative refinement algorithm:

\begin{equation}
\begin{aligned}
&\textbf{Algorithm: Multi-Constraint KKT Solution} \\
&\text{1. Initialize: } \Delta\theta^{(l)} = \mathbf{0} \\
&\text{2. For } k = 1, 2, \ldots, K_{\max}: \\
&\quad \text{a. Update current state: } \theta_{\text{current}}^{(l)} = \theta^{(l)} + \Delta\theta^{(l)} \\
&\quad \text{b. Compute residuals: } r_i^{(k)} = \theta_{\text{current}}^{(l)} h_{\text{intermediate}} - c_i \\
&\quad \text{c. Find most violated: } j^* = \arg\min_i \|r_i^{(k)}\|_2 \text{ s.t. } \|r_i^{(k)}\|_2 < \epsilon \\
&\quad \text{d. If no violations, terminate} \\
&\quad \text{e. Compute single-constraint solution for } c_{j^*} \\
&\quad \text{f. Apply damped update: } \Delta\theta^{(l)} \leftarrow \Delta\theta^{(l)} + \alpha \Delta\theta_{j^*}^{(l)} \\
&\text{3. Return } \Delta\theta^{(l)*}
\end{aligned}
\end{equation}


\end{document}